\documentclass[final]{cvpr}

\usepackage{times}
\usepackage{epsfig}
\usepackage{graphicx}
\usepackage{amsmath}
\usepackage{amssymb}
\usepackage{multirow}
\usepackage{subfigure}
\usepackage{graphicx}
\usepackage{booktabs} 
\usepackage{setspace}
\usepackage{caption}
\usepackage{float}
\usepackage{url}
\usepackage[pagebackref=true,breaklinks=true,colorlinks,bookmarks=false]{hyperref}

\def\cvprPaperID{8875} % *** Enter the CVPR Paper ID here
\def\confYear{CVPR 2021}

\begin{document}
% \setlength{\baselineskip}{22pt}

%%%%%%%%% TITLE
\title{Reducing the Teacher-Student Gap via Spherical Knowledge Disitllation}

\author{%
Jia Guo \\
Zhejiang University\\
{\tt\small jiuzhou@zju.edu.cn}
\and
Minghao Chen \\
  Zhejiang University \\
  {\tt\small minghaochen01@gmail.com}
  \and
Yao Hu \\
  Alibaba Group \\
  {\tt\small yaoohu@alibaba-inc.com}
  \and
Chen Zhu \\
  Alibaba  Group  \\
  {\tt\small none.zc@alibaba-inc.com}
\and
Xiaofei He \\
  Zhejiang University \\
  {\tt\small xiaofei\_h@qq.com}
  \and
Deng Cai \\
  Zhejiang University \\
  {\tt\small dengcai78@qq.com}
% \and
% Yao Hu \\
% Alibaba Group \\
% \t\small{none.zc@alibaba-inc.com}
}

% \author{
%   Jia Guo \\
%   Zhejiang University\\
%   \tt\small{jiuzhou@zju.edu.cn} 
%   \and
%   Minghao Chen \\
%   Zhejiang University \\
%   \tt\small{minghaochen01@gmail.com}
%   \and
%   Yao Hu \\
%   Alibaba Group \\
%   \tt\small{yaoohu@alibaba-inc.com}
%   \and
%   Chen Zhu \\
%   Alibaba  Group \\
%   \tt\small{none.zc@alibaba-inc.com}
%   \and
%   Xiaofei He \\
%   Zhejiang University \\
%   \tt\small{xiaofei_h@qq.com}
%   \and
%   Deng Cai \\
%   Zhejiang University \\
%   \tt\small{dengcai78@qq.com}
% }

\maketitle

%%%%%%%%% ABSTRACT
\begin{abstract}
% Knowledge distillation (KD) aims at obtaining a compact and effective model (student) by learning the mapping function from a much larger one (teacher). 

% A problem with KD is that when trained by a larger teacher, the \emph{student}  performance would unexpectedly drops, such problem is due to the gap between the \emph{student}  and teacher. High capacity network output tend to be overconfident and We find that \emph{student}  struggles to learn this overconfidence. We propose to mitigate this gap of confidence by normalizing the logits before final softmax. 

% However, it is found that a gap between \emph{teacher} and \emph{student}  

% capacity gap
% confidence
% norm
% skd

% We find that the confidence does need to be transferred, and could harm the \emph{student}  performance if forces the \emph{student}  to learn this confidence. We propose Spherical Knowledge Distillation to eliminate this gap explicitly that eases the underfitting problem. We find this novel knowledge representation can train compact models robustly, both to the \emph{teacher} capacity and temperature. The latter, is a hyperparameter of distillation, that can be explained as to *** capacity gap problem by our new view.
% We conducted experiments on both CIFAR100 and ImageNet, and achieve significantly improvement. Specifically, we train ResNet18 to 73.0 accuracy, which is a great improvement over previous SOTAs, and is on par with resnet34 that almost twice the size of \emph{student}.

Knowledge distillation aims at obtaining a compact and effective model by learning the mapping function from a much
larger one. Due to the limited capacity of the student, the \emph{student}  would underfit the teacher.  Therefore, \emph{student}  performance would unexpectedly drop when distilling from an oversized teacher, termed the capacity gap problem.
We investigate this problem by study the gap of confidence between \emph{teacher} and \emph{student}.
We find that the magnitude of confidence is not necessary for knowledge distillation and could harm the \emph{student}  performance if the \emph{student}  are forced to learn confidence.
We propose Spherical Knowledge Distillation to eliminate this gap explicitly, which eases the underfitting problem. We find this novel knowledge representation can improve compact models with much larger teachers and is robust to temperature. We conducted experiments on both CIFAR100 and ImageNet, and achieve significant improvement. Specifically, we train ResNet18 to 73.0 accuracy, which is a substantial improvement over previous SOTA and is on par with resnet34 almost twice the \emph{student} size. The implementation has been shared at https://github.com/forjiuzhou/Spherical-Knowledge-Distillation.

\end{abstract}

%%%%%%%%% BODY TEXT
\section{Introduction}
Deep neural networks have achieved remarkable success in many fields, 
but these state-of-the-art models are often computationally expensive and memory intensive, which hinders their deployment in practice.
Therefore, model compression has become a hot topic and attracted much attention. In 2015, knowledge distillation~(KD) was proposed by~\cite{HintonKD}, where a smaller model~(i.e., \emph{student}) was trained by a larger and well-trained model~(i.e., \emph{teacher}) to achieve model compression. And as a model-agnostic method, it has quickly become a main-stream neural network compression way and inspired previous works.~\cite{FitNets, CRD, SimilarityKD, AttnKD}.

% 1. capacity gap problem
% 2. underfiting 
% 3. overconfidence
% 4. 发现问题
% 5. 解决。
% network 包含多少参数，由目标函数所引导，

The intuition behind KD is to let \emph{student} learn the mapping function of \emph{teacher} to learn its generalization. Naturally, one would expect to train a better \emph{student} with a large and accurate teacher. However, previous researches have invalidated this hypothesize. The \emph{student} performance drops, and the KD loss increases with an oversized teacher, indicating the \emph{student} is unable to mimic an oversized teacher\cite{Earlystop, mirzadeh2019improved, RouteKD}. 
Different capacity models could have disagreements on how to model the data distribution.
Considering that \emph{students} would have insufficient parameters to fit teachers, which usually consists of millions of redundant parameters, it may not surprise that \emph{students} underfit \emph{teachers} eventually.  

% It is widely known that some complex mapping functions are beyond reach for certain neural networks with limited capacity. For example, the well known XOR problem is impossible for a perceptron to fit.

% A broadly concerned problem related to this is that if  \emph{teachers} are bigger and more accurate, \emph{student}  performance would drop .

\begin{figure}[t]
\begin{center}
   \includegraphics[width=0.95\linewidth]{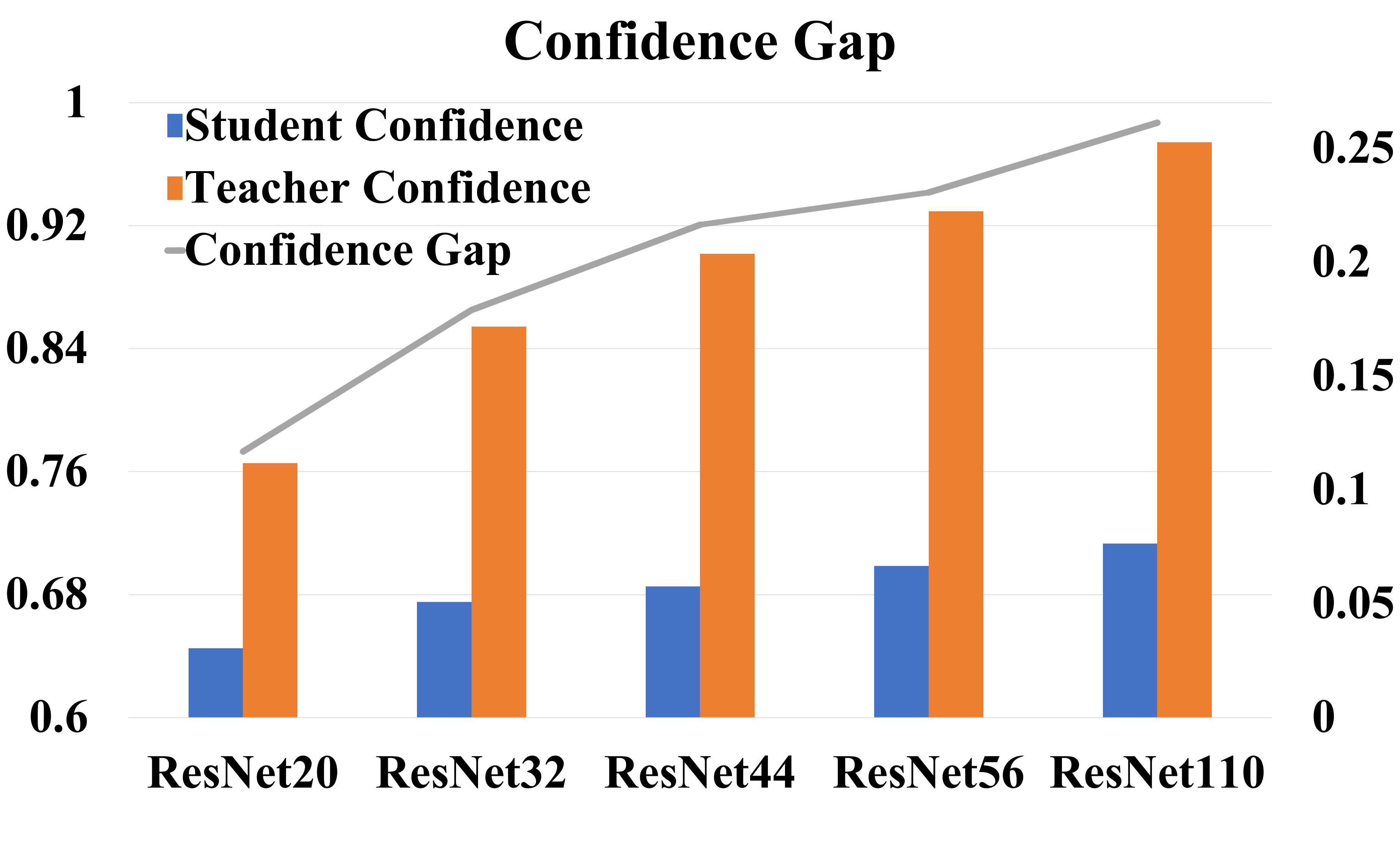}
\end{center}
\vspace{-0.6cm}
   \caption{Confidence Gap Elaboration. When the same \emph{student}  (ResNet14) distilled by larger teacher, the confidence gap continues to increase. }
\label{fig:confidence_gap}
\end{figure}

% For example,  considering the 0/1 loss is indifferentiable.
% 拆解
% 
% the solution the large \emph{teacher} has found is simply not in the solution space of the small student.
We argue that the gap of confidence (i.e., the predicting probability) between  \emph{teachers} and \emph{students} may cause this underfitting problem. Modern neural networks tend to be overconfident, which produces too large predicting probabilities, especially for those large-capacity networks \cite{guo2017calibration,lee2018training, pleiss2017fairness}.
The reason is that larger capacity networks could manage to further minimize negative log likelihood~(NLL) by increasing its confidence, even when the model can correctly classify (almost) all training samples. 
However, compact models may have difficulty fitting this overconfidence. As shown in Fig.~\ref{fig:confidence_gap}, the \emph{teacher} and \emph{student}  confidence difference becomes larger and larger when more expressive networks is distilled. 
When training compact models, it might harm the \emph{student}  performance to force the \emph{student}  to fit the \emph{teacher} overconfidence. 

% However, considering that forcing the \emph{student}  to learn the \emph{teacher} overconfidence might harm its generalization. do we need to learn this \emph{teacher} overconfidence, to train accurate compact models? 

% When training models by minimizing the negative log likelihood loss (NLL), these models tend to be overconfident, especially for those modern large capacity networks. 
% deteriorate \emph{student}  generalization. 

% One principle to train a network is that, the objective function used for training should reflect the true objective as closely as possible. 
% After all, confidence is irrelevant when it comes to predict the right answer, as long as the model correctly assign the largest probability.

% When training the teacher, one should minimize the 0/1 loss for the classification problem. Though in practice, the common method is to use the negative log likelihood loss(NLL) as the proxy objective. 

% Similar phenomenon could also happens to these compact models we want to optimize, as to fitting the peaky distribution of the cumbersome models. 

% \begin{figure}[t]
% \begin{center}
% % \fbox{\rule{0pt}{2in} \rule{0.9\linewidth}{0pt}}
%   \includegraphics[width=0.95\linewidth]{pic/confidence_gap_highres.png}
% \end{center}
%   \caption{Confidence Gap Elaboration. When the same \emph{student}  (ResNet14) distilled by larger teachers, the confidence gap continues to increase. }
% \label{fig:confidence_gap}
% \end{figure}

% \begin{figure}[t]
% \begin{center}
% % \fbox{\rule{0pt}{2in} \rule{0.9\linewidth}{0pt}}
%   \includegraphics[width=0.95\linewidth]{pic/confidence.png}
% \end{center}
%   \caption{We can find that a large gap between the \emph{student}  and \emph{teacher} comes from the confidence difference. If we could rescale the \emph{student}  norm to the same magnitude to the teacher, the gap of output distribution would be reduced by a large margin.}
% \label{fig:confidence_norm}
% \end{figure}

% As \cite{} point out, 大模型有着overconfidence的倾向。
% 小模型很难拟合出 high confidence

% for example, to fit the discrete distribution of the training set. As \cite{} points out, the 

% However, if one want to transfer the knowledge of this trained model to another one

% Normally, NLL can train the model well. However, when the model is able to correctly classify (almost) all training samples, NLL can be further
% minimized by increasing the confidence of predictions. 

% To this end, it could be unfeasible for a small network to fit the sharp distribution of the larger one. 

% 一个naive temperature，not work
% we further test other 

% we found that norm relate to this 
% normlaized logits can help this

Considering most neural networks use softmax function to scale its original output vector~(i.e., logits) to generate normalized probability distribution as its final prediction result, the norm of logits highly determines the output confidence. 
% 换个说法
We investigate this gap of confidence by the norm of logits, and we demonstrate that this confidence information would interfere with the knowledge distillation process. Through extensive experiments, we conclude that it is better not to learn the magnitude of confidence. To this end, we propose Spherical Knowledge Distillation, which projects both the \emph{teacher} and \emph{student}  onto the same sphere by normalizing their logits. SKD highly reduces the gap between  \emph{teachers} and \emph{students} to train efficient and effective students, even when given a much larger teacher.
% Finally, we can get well-generalized knowledge representation, that reduce the capacity gap between the \emph{teacher} and student.

Furthermore, we also highlight another advantage of our SKD: 
KD is sensitive to temperature, which is a very important hyperparameter in distillation. Instead, in our SKD, the \emph{student}  and \emph{teacher} logits are normalized by its own magnitude, so there is less need for using temperature to adjust the softness.
In the experiments, we verify that the performance of our method is very robust to the temperature.
% 斟酌下

We present comprehensive experiments on CIFAR-100 and ImageNet datasets to evaluate our method. We show that: 1) Our SKD mitigates the capacity gap problem considerably, producing results substantially better than previous SOTA. 
2) Our SKD is easy to optimize, exhibits much lower training/test loss with oversized  \emph{teachers} than KD.
3) Our SKD is much more robust with temperature than Hinton distillation.
% To the present, KD is used to distill extensive networks, e.g., Bert. This advantage of SKD could save much time to tune an appropriate temperature.

On the ImageNet classification dataset\cite{ImageNet}, we are the first to obtain excellent results distilled by a very large \emph{teacher} up to ResNet152. Our Resnet-18 distilled by ResNet-50 obtain 73.0\% accuracy that is on par with ResNet34 (73.3\%), which is far better than previous SOTA. 

The rest of the paper is organized as follows: We first analyze the capacity gap problem from the perspective of model confidence.
After that, with some experiments and analysis, we propose the spherical knowledge distillation. Then we will demonstrate that SKD is robust to capacity gap and temperature, and shows excellent performance on CIFAR100 and ImageNet.

\section{Related Work}
\textbf{Knowledge Distillation:}
Buciluǎ et al.~\cite{ModelCompression} first proposed compressing a trained cumbersome model into a smaller model by matching the logits between them. ~\cite{HintonKD} advanced this idea and formed a more widely used framework known as knowledge distillation (KD). Hinton KD tries to minimize the KL divergence between the soft output probabilities generated by the logits through softmax. 
In the later works, the main direction to improve the knowledge distillation is to transfer more knowledge from the \emph{teacher} to the students, such as the intermediate representation~\cite{FitNets, AttnKD} and the relation among instances~\cite{CorrelationKD, GraphKD, SimilarityKD}. FitNets~\cite{FitNets} uses the teacher's intermedia feature map as the additional training hints for the student. The attention transfer method~\cite{AttnKD} makes the activation-based and gradient-based spatial attention map of the \emph{student}  consistent with the teacher. Peng et al.~\cite{CorrelationKD} transfers only the instance-level knowledge and the correlation matrix between instances.  Similarity-preserving KD~\cite{SimilarityKD} requires the \emph{student}  to mimic the pairwise similarity map between the instance features of the teacher. However, there are few works study and modify the origin knowledge distillation loss~\cite{ModelCompression, HintonKD}.

\textbf{Capacity Gap:}
Despite the success of distillation, a mysterious fact found by ~\cite{Earlystop, RouteKD, mirzadeh2019improved} is, a larger \emph{teacher} often harms the distillation despite its more powerful ability, which is called the capacity gap problem.
This problem is particularly severe on ImageNet, resulting in poor performance for KD.
It was widely accepted that the mismatch of capacity caused this problem. 
Due to the lack of understanding of the capacity gap's dynamics, the previous research proposed reducing the teacher's capacity to alleviate this problem heuristically. 
~\cite{Earlystop} propose stopping the training of the \emph{teacher} early. Inspired by curriculum learning, ~\cite{RouteKD} selects several checkpoints generated during the training of the \emph{teacher} and makes the \emph{student}  gradually mimic those checkpoints of the \emph{teacher}. And ~\cite{mirzadeh2019improved} propose to use a medium-size \emph{teacher assistant} (TS), to perform a sort of sequence distillation, that is, the TS first learn from the teacher, then the \emph{student}  can learn from this TS. By directly reduce the capacity of teachers, these methods also scarify the \emph{teacher} performance.
% We find that the norm of logits is the poisonous part of the teachers' output that caused this problem. By addressing this norm problem, we can achieve much better performance without sacrifice teachers performance.
We investigate the gap of confidence between the different sizes of models. We find that the spherical knowledge representation is a more suitable objective function for a compact model to learn. 

\textbf{Normalization:}
The methodology of SKD correlates to the normalization method. Normalization is widely used in fields of deep learning. For example, batch normalization~\cite{ioffe2015batch} and layer normalization~\cite{ba2016layer} are used in current SOTA computer vision and natural language processing models. Within the knowledge distillation field, \cite{AttnKD} applies normalization when distilling the teacher's feature map. The normalization is used to form an "attention map" in the feature layer.  
Face recognition \cite{SphereFace, CosFace} also use normalization on feature vector to compute the cosine similarity between two input. 

Unlike the previous works that normalize the feature map or the weight, we normalize the logits specially designed for knowledge distillation, to reduce the gap between \emph{teacher} and student. To the best we know, we are the first to apply the normalization before the last softmax.

% \textbf{Model Confidence}
% In real-world decision-making systems, classification networks must be accurate and indicate when they are likely to be incorrect. Specifically, a network should provide a calibrated confidence measure in addition to its prediction. In other words, the probability associated with the predicted class label should reflect its ground truth correctness likelihood\cite{guo2017calibration}.

% While neural networks today are undoubtedly more accurate than they were a decade ago, they tend to be overconfident. That is to say, modern deep neural networks tend to predict with a very high probability compared to its accuracy. This phenomenon is more severe on bigger networks.
\section{Teacher-Student Gap}
\subsection{Hinton Knowledge Distillation}
From an abstract perspective, a neural network learned during training is the mapping function from input X to output vector. And as for a well-trained model, besides the mapping from X to the corresponding correct answer, the probabilities assigned to those incorrect answers can also unveil some knowledge. For example, as for a picture of ``cat'', although it seems impossible to predict it as `dog'', this mistake is still many times more likely than mistaking it for ``air plane''. Such a correlation between ``cat'' and ``dog'' contained in the output vector implies the relationship between the two classes and unveils how a model tens to generalize. 
Following this intuition, Hinton et al~\cite{HintonKD} proposed a model compression method, named Knowledge Distillation, where compress knowledge of larger model~(i.e.~\emph{teacher}) into a smaller model~(i.e.~\emph{student}) by using the class probabilities predicted by~\emph{teacher} as ``soft target'' of~\emph{student} when training. Let us define $f^{T}(x)$ is the logits vector of~\emph{teacher} for sample $x$ and $f^{S}(x)$ is the logits vector of~\emph{student}. $y$ is the label vector. The loss of KD when training~\emph{student} can be defined as:
% , namely ``knowledge'', actually
\begin{align}
\mathcal{L} = \lambda \mathcal{L}_{KD} + (1-\lambda) \mathcal{L}_{cls} \\
\mathcal{L}_{KD} = \sum_{i}p_i^T\log p_i^S \\
\mathcal{L}_{cls} = \sum_{i}y_i\log p_i^S&, 
% \mathcal{\rm Under certain approximation} \\ {L}_{MSE} = &\sum_{i}(f_i^S(x)-f_i^T(x))^2 
\end{align}
where $p_i = \frac{exp(f_i(x)/\tau)}{\sum_j exp(f_j(x)/\tau)}$. 
We can find $\mathcal{L}_{cls}$ is typical cross entropy loss. $\mathcal{L}_{KD}$ is the cross entropy loss for fitting ``soft target'', and $\tau$ is a hyper-parameter, named temperature, to control its ``softness''. The final loss is the sum of these two losses weighted by trade-off parameter $\lambda$.

\subsection{Capacity Gap Problem}
Since the idea of KD is transferring the knowledge of \emph{teacher} into \emph{student}, we naturally think a larger and more accurate \emph{teacher} can capture more knowledge and thus can serve as better supervision to \emph{student} in KD. Unfortunately, some experiment results invalidate this hypothesize~\cite{Earlystop,mirzadeh2019improved}, that with a larger teacher, the performance of \emph{students} would degenerate unexpectedly. We term this problem as Capacity Gap.

\begin{figure}[ht]
\begin{center}
   \includegraphics[width=0.99\linewidth]{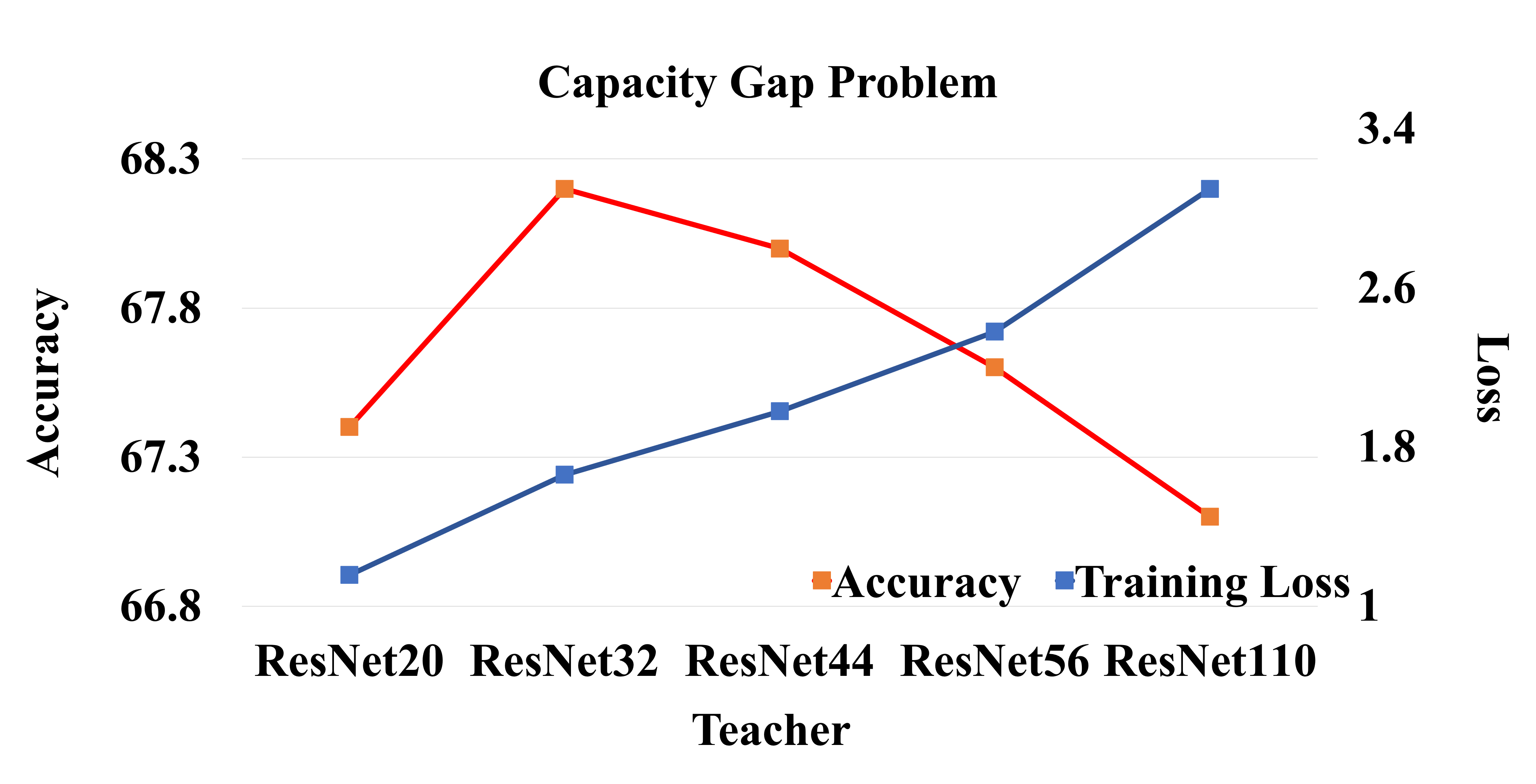}
\end{center}
   \caption{The capacity gap problem. 1) The ResNet14 \emph{student}  accuracy drops when the \emph{teacher} gets bigger after ResNet32. 2) The KD loss continues to increase, indicating that the \emph{student}  underfits the teacher.}
\label{fig:capacity_gap}
\end{figure}

To dig deeper into this ``capacity gap'', we observe the KD loss between \emph{teacher} and student. From Fig.\ref{fig:capacity_gap}, as the \emph{teacher} grows, the KD loss increase, showing more discrepancies with the teacher. This phenomenon indicates that the \emph{student}  does not fit the \emph{teacher} output objection well enough.
In other words, under the insufficient capacity, \emph{students} would underfit the mapping function of teachers, especially when the gap is vast. 
%Due to the limited capacity, \emph{students} are indeed prone to underfitting.

This problem, in our perspective, stems from the way we define knowledge. In the following sections, we find that the original probability distribution is an inappropriate objective.
Then we explore a new form of knowledge representation, which excludes the difficult factor for \emph{students} to fit and does not serve the model generalization, i.e., the model confidence.

% \textbf{The Bias of the teacher}
\subsection{Model Confidence Gap}

It has been known that when a neural network becomes bigger,
along with the increase of model capacity, neural networks
often tend to become more confident and assign a higher
probability to the predicted label. This phenomenon has attracted some attentions \cite{guo2017calibration} and provides us a new perspective to study the capacity gap problem in KD. For example,in \cite{guo2017calibration}, model confidence is quantified by the probability
assigned to the predicted label. Chuan et al. demonstrate
the significant positive relationship between the model confidence and the model size, and try to explain it by negative
log likelihood (NLL), the standard loss of neural network.

As we knew, NLL is minimized if and only if the model
output totally recovers the ground truth, which is the onehot encoding of the true label in classification. Thus, during a neural network training, even after it can classify (almost) all training samples correctly, NLL can be further
minimized by assigning more probability to the true label.
Because a larger model (e.g., teacher in KD) is often more
powerful and can further lower NLL, it tends to be more
confident and assign a higher probability to the predicted
label than a smaller model.

During the training of KD, student directly mimics the
“soft target” generated by teacher, so it inevitably mimics
the confidence of teacher together, which is actually highly
related with model size and thus is hardly transferable between two models. We think the capacity gap problem is
probably due to the original KD does not eliminate the influence of the confidence gap between teacher and student.
Following this assumption, we redefine the “soft target” by
normalization to calibrate model confidence and further alleviate the confidence gap. Next, we would introduce our
solution in detail.

\begin{table*}[]
\begin{center}
\begin{tabular}{llllllll}
\toprule
\begin{tabular}[c]{@{}l@{}}Teacher\\ Student\end{tabular} & \begin{tabular}[c]{@{}l@{}}WRN-40-2\\ WRN-16-2\end{tabular} & \begin{tabular}[c]{@{}l@{}}ResNet56\\ ResNet20\end{tabular} & \begin{tabular}[c]{@{}l@{}}ResNet110\\ ResNet32\end{tabular} & \begin{tabular}[c]{@{}l@{}}vgg13\\ vgg8\end{tabular}  & \begin{tabular}[c]{@{}l@{}}DenseNet-112-33\\ DenseNet-40-20\end{tabular} \\\midrule
Teacher & 75.61  & 74.31 & 74.31  & 74.64  & \quad 81.68 \\
Student & 73.26  & 69.50 & 71.14 & 70.36  & \quad 74.32 \\
Norm loss & 72.95  & 69.34 & 70.56  & 70.41 & \quad 74.10 \\
KD & 74.92& 70.7 & 73.08 & 72.98   & \quad 75.50 \\
KD$*$ & 75.21 & 71.2 & 73.47  & 73.43 & \quad  76.34  \\
SKD & \textbf{75.75}  & \textbf{72.08} & \textbf{74.09}  & \textbf{74.17} & \quad \textbf{77.70}
\\\bottomrule
\end{tabular}
\end{center}
\caption{Evaluating the effectiveness of norm and normalized logits on CIFAR-100 dataset. ``Norm loss'' means \emph{student}  learned from the norm of the teacher. `$*$' means KD with teacher's logits normalized. We can find  that KD by normalized \emph{teacher} logits achieve better performance than original KD.}
\label{normloss}
\end{table*}

\section{Spherical Knowledge Distillation}
In this section, we will first explain our assumption about how to alleviate the influence of the confidence gap in KD and conduct two experiments to support it. After analyzing the gradients to the logits, we will introduce our proposed KD method, named Sphere Knowledge Distillation~(SKD), in detail.

\subsection{Preliminary}\label{norm_and_normalized_logits}
We follow the idea in~\cite{guo2017calibration} and quantified model confidence by its probability of the predicted label. Because most nerual network pass the output logits through softmax function to generate the probability distribution, the peak value (confidence) of the final prediction is highly determined by the magnitude of logits. Thus, compared with logits, normalized logits would be less affected by model confidence, but still keep the correlation among classes. Therefore, we think normalized logits would be a better knowledge representation for knowledge distillation.
To validate the above assumption, we did two experiments to analyze the effectiveness of norm and normalized logits in KD, respectively. Mathematically, we decompose the logits of~\emph{student} as 
$f^S(x) = l^S(x) * \hat f_i^S(x)$
where the $l^S(x)$ is the $l^2$ norm, defined as
$l^S(x) = (\sum_i {f_i^S(x)}^2)^{\frac{1}{2}}$, and $\hat f_i^S(x)$ is the normalized logits:
$\hat f_i^S(x) = f_i^S(x)/l^S$. The \emph{teacher} can be defined in the same way. 
% Therefore, larger teachers will not bring additional performance benefits to students.

% To that end, what one can do to reduce this gap? The previous method is to put highly regulation on the teacher, like early stopping the training of teacher, to explicitly narrow the gap between \emph{student}  and teacher. But such approach would also affects the \emph{teacher} performance.
% In this section, we investigate the role of norm during the distillation. In specific, by dissecting the loigts into two parts, norm and normalized logits.
% Based on the above analysis, we find that both both norm and the normalized logits can be used to measure the gap between \emph{students} and teachers.
% When the \emph{teacher} capacity grows, the performance of \emph{teacher} and this gap enlarges together. 

% So a question naturally arises, does the \emph{student}  need to learn both these two values? especially when these two loss are equally hard. Intuitively, the normalized logits can contain much more information than the length of a logits. We assume that the relative size of values in logits is sufficient to represent knowledge, and norm is unnecessary information. To verify this hypothesis, we designed two experiments. 
% First, we obtained two baseline model that training with KD or ground truth respectively. 
% We first train a baseline model using ResNet20 as \emph{student}  and ResNet56 as teacher, on CIFAR100 dataset.

% \textbf{Directly Learn norm of The \emph{teacher} Experiment}
\textbf{Evaluating the Effectiveness of Norm of Logits in KD.} \label{normalized_logits}
To evaluate it, here we train a \emph{student}  by replacing the standard $\mathcal{L}_{KD}$ with the MSE loss of the logits norm between \emph{student} and \emph{teacher}:
$
\mathcal{L} = \lambda ||l^S(x)-l^T(x)||_2 + (1-\lambda) \mathcal{L}_{cls}
$
The results are shown in Table~\ref{normloss}. We can find its accuracy~(“Norm loss”) is even worse than that training with only ground truth. It indicates that transferring the norm of logits would not benefit~\emph{student} and might be even harmful.

% here we distilled a ResNet20 as \emph{student} from a well-trained ResNet56 by 

% This experiment was meant to examine if the transferring of the norm would bring improvement to \emph{student}  performance. The \emph{student}  is trained to minimize the MSE loss of $l^{2}$ logits norm between the \emph{student}  and the teacher, along with the cross-entropy loss with ground truth. 

% The accuracy of this experiment is worse than that training with only ground truth, which indicates that directly transferring the norm from \emph{teacher} to \emph{student}  is harmful to the student.

% \textbf{Learn Normalized Logits From \emph{teacher} Experiment}
\textbf{Evaluating the Effectiveness of Normalized Logits in KD.}
Here we try to replace the standard $\mathcal{L}_{KD}$ with a norm-independent loss. Specifically, we first normalize logits of~\emph{teacher} and then multiply it by its average $l^2$ norm as the ``soft target'' in training~\emph{student}. We select a wide range of \emph{teacher} and \emph{student}  settings. The results are shown in Table~\ref{normloss}, denoted as KD*. We can find its accuracy is higher than that of the Hinton distillation. This improvement implies that compared with original logits, the normalized logits is probably better for transferring knowledge between~\emph{student} and~\emph{teacher}.

% This experiment is conducted by normalizing the teacher's logits and then multiply it by its average $l^2$ norm. Other settings are consistent with the above experiment. In this way, all the teacher's logits will be projected onto a sphere, and all information contained in the \emph{teacher} $l^2$ norm would be eliminated. 

% We can see that the accuracy is improved than the Hinton KD. When the \emph{teacher} fit the distribution of the dataset, the norm can be rather complicated. Even \emph{teacher} can classify all examples into right categories, it still could maximize the NLL by enlarge the norm of some examples. These norm information could be rather nasty. In this experiment, \emph{students} do not have to learn the complicated distribution of teacher's $l^2$ norm among the training dataset. The improved accuracy demonstrates that norms of \emph{teacher} logits actually harm the distilling process.

\textbf{Norms Affect the Learning of Normalized Logits.}
Next, we will further explore the observation of the above two experiments from a gradient perspective. 
We found that in Hinton KD, the normalized logit learning will be interfered with by the norm. Thus distillation efficiency can be highly affected.

For simplicity, we conduct from the MSE loss of logits, which is a special case of Hinton distillation\cite{HintonKD}.
We compute the gradient on $\hat f_i^S(x)$ as follows:
\begin{equation}
\label{equ:inequal}
\begin{aligned}
     \frac{\partial \mathcal{L}_{KD}}{\partial \hat f_i^S(x)} &=\partial ||l^S(x) \hat f^S(x)-l^T(x) \hat f^T(x)||_2 / \partial \hat f_i^S(x)\\
     &=2l^S(x)(l^S(x) \hat f_i^S(x)- l^T(x)\hat f_i^T(x)) \\
\end{aligned}
\end{equation}
We can see that while the \emph{student}  tries to learn the $\hat f_i^T(x)$, the gradient of $\hat f_i^S(x)$ is interfered with by both \emph{teacher} and \emph{student}  norms, and it is worth noting that the latter is changing all the time during the distillation process.  Therefore, in the distillation process, $\hat f_i^S(x)$ would be difficult to converge to $\hat f_i^T(x)$, which is more useful knowledge than logits. 

After normalizing the \emph{teacher} logits as in "KD*", the \emph{teacher} norm would be a constant, however, that still leaves the \emph{student}  norm to interfere with $\hat f_i(x)$. If we also normalize the \emph{student}  logits, the optimization to $\hat f_i(x)$ would be free from the norm $l^S(x)$ and be much easier.

% To see how \emph{student}  norm changes in distillation, we compute the gradient on the norm:
% \begin{equation}
% \label{equ:inequal}
% \begin{aligned}
%      \frac{\partial \mathcal{L}_{MSE}}{\partial l^S} &=2\sum_{i}(l^S f_i^S(x)- l^T f_i^T(x))\hat f_i^S(x)\\
%      %&=2\sum_{i}l^S{\hat f_i^S(x)}^2-l^T\hat f_i^T(x)\hat f_i^S(x) \\
%      &= 2 (l^S-l^T \sum_{i} \hat f_i^T(x)\hat f_i^S(x))
% \end{aligned}
% \end{equation}

% We can see that \emph{student}  tries to learn the \emph{teacher} norm, but again, this process would also be interfered by the $\hat f_i(x)$. These two terms are interfering with each other that could harm the stability of the distillation.

% We can see that \emph{students} are trying to learn \emph{teacher} norms, but again, this process will also be interfered by normalized logits. 

% \textbf{Observation 1}  Norm and normalized logits would interfering with each other, that could harm the stability of the distillation.
% 在传统的kd中，norm会干扰对于真正知识的学习。
% Due this capacity gap problem, the original probability distribution is not a good objective for the \emph{student}  to learn.

\begin{figure*}[t]
\begin{center}
   \includegraphics[width=0.93\linewidth]{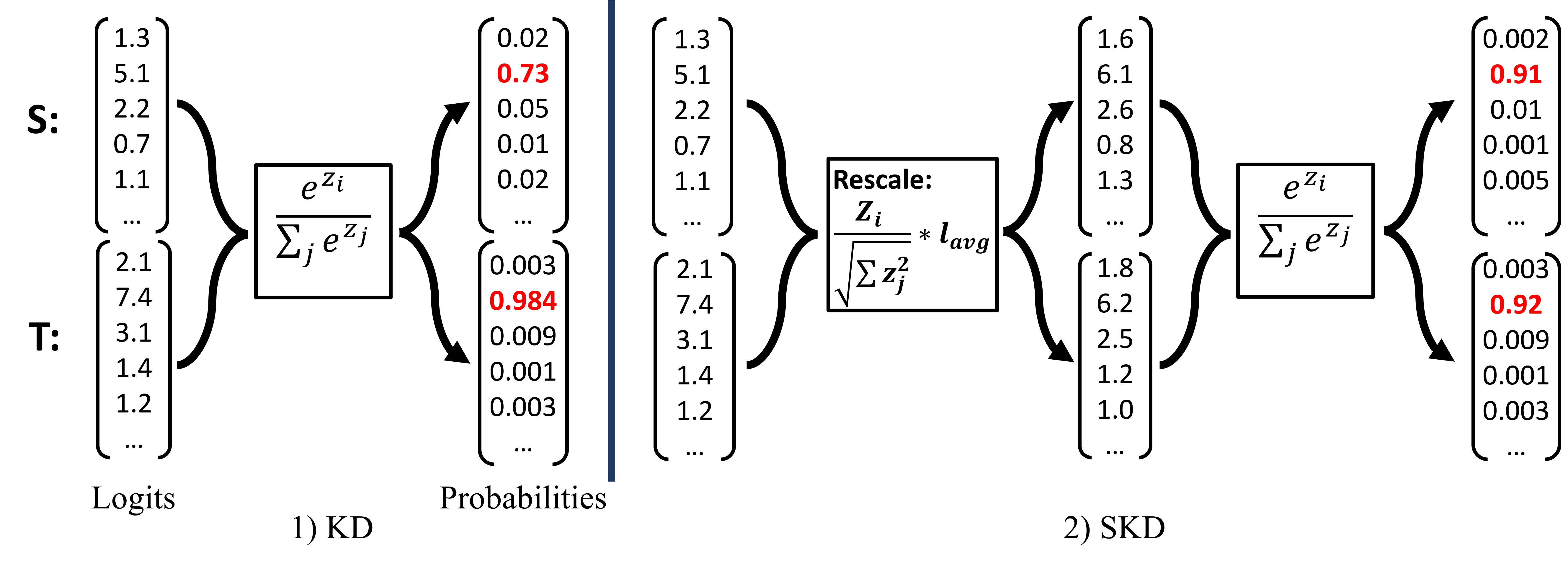}
\end{center}
   \caption{ 1) The \emph{teacher} has large norm logits and predicts high confidence, while the \emph{student}  with small norm logits predicts low confidence. KD directly mimics original \emph{teacher} output, which would force the \emph{student}  to learn \emph{teacher} overconfidence. 
   2) SKD rescales \emph{teacher} and \emph{student}  logits to the same magnitude, thus the teacher-student confidence gap would be reduced by a large margin.}
   
%   We can find that a large gap between the \emph{student}  and \emph{teacher} comes from the norm of logits. If we could }
\label{fig:SKD}
\end{figure*}

\subsection{Spherical Knowledge Distillation}

% \begin{figure}[htbp]
% \centering
% \begin{minipage}[t]{0.45\linewidth}
% \centering
% \includegraphics[width=3.5cm]{pic/raw.png}
% \end{minipage}
% \begin{minipage}[t]{0.45\linewidth}
% \centering
% \includegraphics[width=3.5cm]{pic/sphere_modified.png}
% \end{minipage}
% \caption{The original distribution of model output differs with each other. With a spherical projection, the representation is more closer}
% \label{sphere}
% \end{figure}

From the above observations, we take that normalized logits contributes to the critical knowledge when distilling knowledge. While the norm of logits mainly forms model confidence, it is relatively useless and could hinder the learning of normalized logits.

Following this paradigm, we propose a more efficient Spherical Knowledge Representation in this matter. Specifically, all logits of the \emph{teacher} and the \emph{student}  would be projected on a sphere by normalization, so the confidence gap is eliminated by a large margin, like Fig. \ref{fig:capacity_acc} showed. Compared with the previous original probability distribution, spherical knowledge considers the \emph{teacher} and \emph{student}  representation bias. Thus it is a more suitable target for the \emph{student} to learn. We provide a visual presentation of SKD on Fig. \ref{fig:SKD}.

The logits $f_i(x)$ first scale to a unit vector $\hat f_i(x)$, then multiply with the average \emph{teacher} norm $l_{avg}$ to recover its norm to the original level, which is important for keeping the supervised learning unaffected.
The spherical knowledge distillation performs as:
\begin{equation}
\begin{split}
\mathcal{L}_{SKD} &=  \sum_{i} \hat p_i^T\log \hat p_i^S,
\\
\mathcal{\rm where \quad} & \hat p_i = \frac{exp(\hat f_i(x)*l_{avg}/\tau)}{\sum_j exp(\hat f_j(x)* l_{avg}/\tau)} 
\\
\mathcal{L}  = \lambda&\mathcal{L}_{SKD} + (1-\lambda)\mathcal{L}_{cls}(\hat p^S,y)
\end{split}
\end{equation}
where $\hat f_i^T(x)$ is the normalized logits, $\lambda$ is the trade-off parameter, $l_{avg}$ is the average norm of the teacher, and $\tau$ is the temperature parameter. Same as Hinton KD, in $L_{cls}$, the temperature would be set to 1.

SKD inherits the simplicity of Hinton KD. The SKD formula only adds a constant $l_{avg}$, which is the new norm. This value is set to the average norm of the teacher. This is because unit vector logits may prevent the confidence from getting close to 1 even when the samples are well trained. In this situation the model will always try to optimize this well-trained example, while the harder examples may not get sufficient training\cite{Wang_2017}.

In experiments, we will demonstrate that SKD can provide excellent performance even with a very large teacher, and is very robust to the temperature.

% One might want to mitigate the gap of confidence by softening the distribution with a large temperature. We conducted extensive experiments in ..., and find that adjusting temperature can not narrow this gap.

% 还可以看成自适应temperautre，norm *m/t
% 我们也尝试了手动调temperature，

% 手动调temperature或许能达成效果，但是会很麻烦。我们在实验部分对这个问题进行了讨论，发现不足以解决这个问题。

% We add the ablation study of the m in the supplement part. To be noticed, this value does not need to be carefully tuned. In practice, m can be set between 2 to 3. Since m and the temperature can be thought as one temperature, and, as we will discussed below, skd is very robust to the temperature. This parameter is for the negative log likelihood loss solely.
% \textbf{the specific value of m}
% SKD multiply the normalized logits with a constant m.
% 

% 1. 画图，球形是什么样德，当温度变化时又是什么样德
% 2. loss下降快
\begin{table*}[!tbp]\centering
\begin{tabular}{llllllll}
\toprule
\begin{tabular}[c]{@{}l@{}}Teacher\\ Student\end{tabular} & \begin{tabular}[c]{@{}l@{}}WRN-40-2\\ WRN-16-2\end{tabular} & \begin{tabular}[c]{@{}l@{}}WRN-40-2\\ WRN-40-1\end{tabular} & \begin{tabular}[c]{@{}l@{}}ResNet56\\ ResNet20\end{tabular} & \begin{tabular}[c]{@{}l@{}}ResNet110\\ ResNet20\end{tabular} & \begin{tabular}[c]{@{}l@{}}ResNet110\\ ResNet32\end{tabular} & \begin{tabular}[c]{@{}l@{}}ResNet32*4\\ ResNet8*4\end{tabular} & \begin{tabular}[c]{@{}l@{}}vgg13\\ vgg8\end{tabular} \\\midrule
Teacher & 75.61 & 75.61 & 72.34 & 74.31 & 74.31 & 79.42 & 74.64 \\
Student & 73.26 & 71.98 & 69.06 & 69.06 & 71.14 & 72.50 & 70.36 \\
KD~\cite{HintonKD} & 74.92 & 73.54 & 70.66 & 70.67 & 73.08 & 73.33 & 72.98 \\
FitNet~\cite{FitNets} & 73.58 & 72.24 & 69.21 & 68.99 & 71.06 & 73.50 & 71.02 \\
AT~\cite{AttnKD} & 74.08 & 72.77 & 70.55 & 70.22 & 72.31 & 73.44 & 71.43 \\
SP~\cite{SimilarityKD} & 73.83 & 72.43 & 69.67 & 70.04 & 72.69 & 72.94 & 72.68 \\
CC~\cite{CorrelationKD} & 73.56 & 72.21 & 69.63 & 69.48 & 71.48 & 72.97 & 70.71 \\
VID~\cite{VID} & 74.11 & 73.30 & 70.38 & 70.16 & 72.61 & 73.09 & 71.23 \\
RKD~\cite{RelationalKD} & 73.35 & 72.22 & 69.61 & 69.25 & 71.82 & 71.90 & 71.48 \\
PKT~\cite{PKT} & 74.54 & 73.45 & 70.34 & 70.25 & 72.61 & 73.64 & 72.88 \\
AB~\cite{AB} & 72.50 & 72.38 & 69.47 & 69.53 & 70.98 & 73.17 & 70.94 \\
FT~\cite{kim2020paraphrasing} & 73.25 & 71.59 & 69.84 & 70.22 & 72.37 & 72.86 & 70.58 \\
FSP~\cite{AGFKD} & 72.91 & - & 69.95 & 70.11 & 71.89 & 72.62 & 70.23 \\
NST~\cite{NST} & 73.68 & 72.24 & 69.60 & 69.53 & 71.96 & 73.30 & 71.53 \\
CRD~\cite{CRD} & 75.48 & 74.14 & 71.16 & 71.46 & 73.48 & 75.51 & 73.94 \\
\textbf{SKD} & \textbf{75.75} & \textbf{75.06} & \textbf{72.08} & \textbf{72.12} & \textbf{74.09} & \textbf{76.40} & \textbf{74.17} \\\bottomrule
\end{tabular}
\caption{CIFAR100 experiments}
% \caption{Performance on CIFAR100 compared with other distillation methods. We highlight the best performance of each setting.
% }
\label{Cifar100}
\end{table*}

\begin{table*}
\begin{center}
\begin{tabular}{lllllllllll}
\toprule
 CE & KD$_{34}$ & ES$_{34}$~\cite{Earlystop} & SP$_{34}$ & CC$_{34}$ & CRD$_{34}$ & AT$_{34}$ & SKD$_{34}$  & SKD$_{34}*$ & SKD$_{50}*$ & SKD$_{101}*$ \\\midrule
69.8 & 69.2 & 71.4 & 70.62 & 69.96 & 71.38 & 70.7 & 72.3  & 72.80 & \textbf{73.01} & 72.85 \\
\bottomrule
\end{tabular}
\end{center}
\caption{Imagenet experiments. All use ResNet18 as the student. The subscript denote the size of the teacher, and $*$ denote training with 100 epochs.}
\label{ImageNet}
\end{table*}

\section{Experiments}
In this section, we will show comprehensive experiment results to validate the effectiveness of SKD from several perspectives. Specifically, we first selected many SOTA KD methods and compared SKD with them on two popular CV datasets to prove its performances. Then we focused on evaluating whether it could alleviate the Capacity Gap problem. Finally, we did some further analysis in terms of parameter robustness and efficiency.

\subsection{The Performance on CIFAR100 \& ImageNet}
Here we conducted experiments on two popular datasets, CIFAR100 and ImageNet. The results well prove the excellent performance of SKD through a broad range of teacher-student settings on CIFAR100 and ImageNet. 

% \textbf{The Training of SKD}
% In figure x, we compare the training/validation errors/losses during the training procedure between KD and SKD. The SKD has lower training error/loss throughout the whole training procedure.
% \\
% \\
% 快，好

\textbf{Baselines.} We selected many SOTA KD methods to evaluate the performances of SKD. 
Knowledge defined from intermediate layers: FitNet~\cite{FitNets}, AT~\cite{AttnKD}, SP~\cite{SimilarityKD}, PKT~\cite{PKT}, FT~\cite{kim2020paraphrasing}, FSP~\cite{AGFKD} .
1) Knowledge defined via mutual information: CC~\cite{CorrelationKD}, VID~\cite{VID}, CRD~\cite{CRD}.
2) Structured Knowledge: RKD~\cite{RelationalKD}.
3) Knowledge from logits: KD~\cite{HintonKD}, NST~\cite{NST}

\subsubsection{Results on CIFAR100} 

\emph{CIFAR100}~\cite{cifar100} is a relatively small data set and is widely used for testing various of deep learning methods. CIFAR100 contains 50,000 images in the training set and 10,000 images in the dev set, divided into 100 fine-grained categories.

\textbf{Experimental settings.}
We run a total of 240 epochs for all methods. The learning rate is initialized as 0.05, then decay by 0.1 every 30 epochs after 150 epochs. Temperature is 4, and the weight of SKD or KD and cross-entropy is 0.9 and 0.1 for all the settings. As for other hyper-parameters of baselines, we follow the setting of~\cite{CRD}.

\textbf{Experiment Results.}
Table~\ref{Cifar100} shows the detailed results, where the first line shows the model of~\emph{teacher}, and the second line shows the model of~\emph{student}. We can find that SKD always has an outstanding improvement compared with all other methods. In some situations~(e.g. those where \emph{teacher}/\emph{student} is WRN-40-2/WRN-40-1 or ResNet110/ResNet32), the performances of SKD are even very closed to those of~\emph{teacher}.
% When the \emph{teacher} and \emph{student} share the same architecture, SKD has a stable improvement compared with all other methods. And when the architecture is different, our method still achieves outstanding performance in most settings. 
% ``It is worth noting that our methods are simpler to implement.'' ???

\subsubsection{Results on ImageNet} 

\emph{ImageNet}~\cite{ImageNet} is a much larger one than CIFAR100. ImageNet contains 1.2M images for training and 50K for validation, that distributes in 1000 classes. 

\textbf{Experimental settings.}
Here we use ResNet18 as~\emph{student} for all of methods. And the initial learning rate is set at 0.1, and decay by 0.1 in 30, 60, 80. Batch size is set as 256. 

% We trained all of baselines by 90 epochs.
% We use 4 NVIDIA 1080ti with distributed learning to accelerate our experiment. 
% For better performance, some experiments use 100 epochs, and others use 90 epochs.  A total 1024 batch size is used. Typically, our SKD with ResNet18 can finish within one day.

\textbf{Experiment Results.}
First, we showed the training process of ResNet18 distilled by ResNet34 in Fig. \ref{fig:test_accuracy}, under both KD and SKD. SKD achieves excellent performance all the time. In the early phase of the 30th epoch, SKD provides comparable performance to KD's final performance.

Table~\ref{ImageNet} shows the detail results. We can find that SKD exceeds all of the previous SOTA by a large margin. Before SKD, the improvement of this task is limited, such as the second highest approach is only 0.02 higher than before (ES[1] compared to CRD[8]).

In contrast, SKD achieved an accuracy of 72.3 under the ResNet34 teacher. Furthermore, We found that the model accuracy still has an upward trend until the end of the training. To further push the limit, we add another ten epochs that push the final result to 72.8 accuracy.

% and apply a strategy that gradually reduces the weight of distillation loss during the training process. This strategy is inspired by \cite{Earlystop}. To be specific, we set $\lambda$ to 0.9 before 60 epoch, and 0.5 before 80 epoch, and 0.1 with the rest of epochs. 

We further conducted distillation with much larger teachers with ResNet50, ResNet101, and ResNet152
\subsection{SKD Alleviates the Capacity Gap Problem}
\begin{figure}[t]
\begin{center}
   \includegraphics[width=0.99\linewidth]{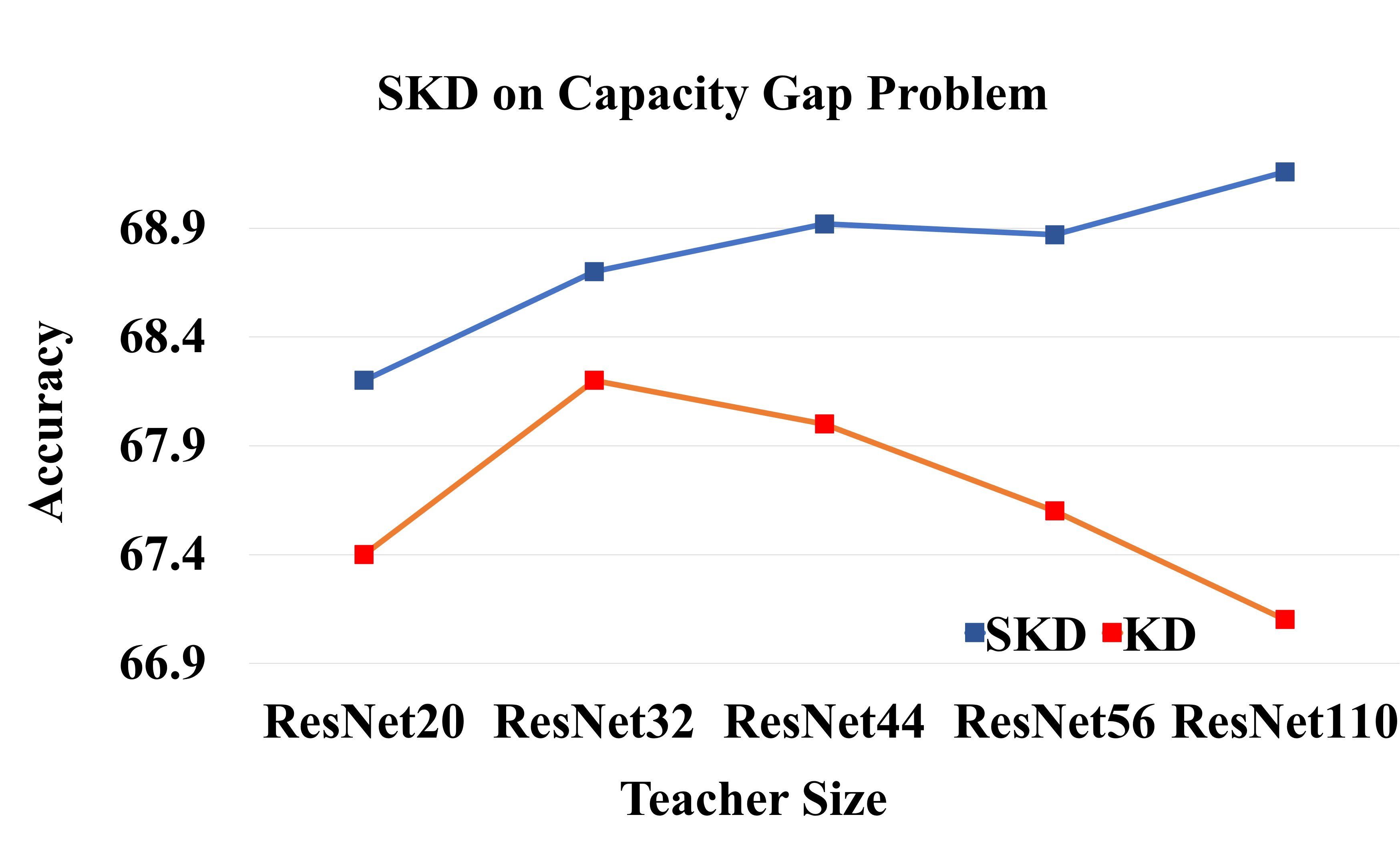}
\end{center}
   \caption{SKD on capacity gap problem with CIFAR-100. SKD provides much better performance than KD with all \emph{teacher} capacities, and can get boosted by bigger teachers.}
\label{fig:capacity_acc}
\end{figure}

SKD highly alleviates the capacity gap problem. Hence, the performance is boosted to an unprecedented level, despite the same limited capacity. On CIFAE-100 task, We trained the ResNet14 with multiple teachers on the CIFAR100 dataset, and the result is shown in Fig. \ref{fig:capacity_acc}. Under SKD, \emph{student} performance continues to increase as the \emph{teacher} gets bigger under this task. 

We also evaluate the capacity gap problem in ImageNet. Some previous methods aim to close this gap by explicitly reducing the \emph{teacher} gap, like Early Stop~\cite{Earlystop} (ES) and \emph{teacher} Assistant~\cite{mirzadeh2019improved} (TA). TA does not provide comparable ImageNet results in their report, while ES offers a similar method with TA in their results, namely the Seq. ESKD. We deliver the above results in Table \ref{Capacitygap}.

We use ResNet18 as the student. We can see that SKD exceeds Early Stop and Seq. ESKD methods with a large margin in all \emph{teacher} settings. For example, when distilled by ResNet50 and ResNet152, the performance exceeds other methods by 2 percents. With ResNet50 as the teacher, we obtained the best performance, 73\% accuracy, which is the first ResNet18 model to our knowledge.

% \subsection{SKD eases the training by focus on learning normalized logits}

\begin{table}[!tbp]
\begin{center}
\begin{tabular}{lllllllllll}
\toprule
 \emph{teacher}  & Method & Accuracy \\
          & KD & 69.43 \\
 ResNet34 & Early Stop & 70.98 \\
          & \textbf{SKD} & \textbf{72.80} \\
\midrule 
          & KD & 69.05 \\
 ResNet50 & Seq. ESKD~\cite{Earlystop} & 70.65 \\
          & Early Stop & 70.95 \\
          & \textbf{SKD} & \textbf{73.01} \\
\midrule 
ResNet101 &  \textbf{SKD} &  \textbf{72.85} \\
\midrule 
          & Seq. ESKD~\cite{Earlystop} & 70.59 \\
ResNet152 & Early Stop &  70.74 \\
          & \textbf{SKD}  & \textbf{72.70} \\
\bottomrule
\end{tabular}
\caption{SKD on capacity gap problem with ImageNet. The baseline is ResNet18, 69.9 accuracy with ground truth. SKD surpass all other methods with a large margin in all \emph{teacher} settings.}
\label{Capacitygap}
\end{center}
\end{table}

\subsection{SKD Eases the Training of Student}
\begin{figure}[h]
\begin{center}
   \includegraphics[width=0.99\linewidth]{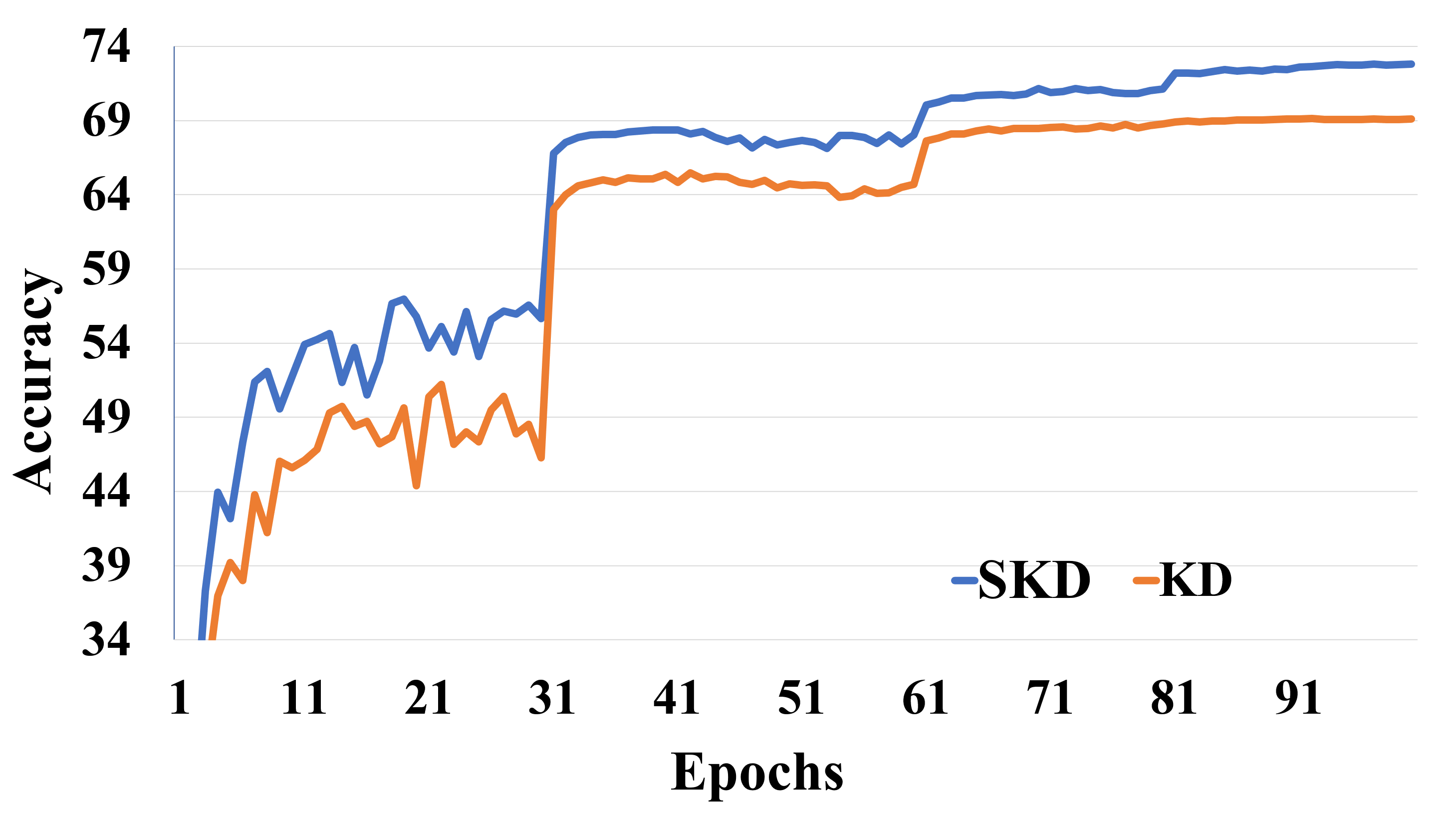}
\end{center}
   \caption{The training process of KD and SKD on ImageNet. SKD achieves comparable accuracy with KD at the 30th epoch.}
\label{fig:test_accuracy}
\end{figure}

Besides mitigating the capacity gap problem, SKD also performs well when distilled with relative small teachers. For example, the ResNet18 distilled by ResNet34 improved nearly 3 points than KD, as shown in Fig. \ref{fig:test_accuracy}. 

% \textbf{The Underfitting Problem is Highly Mitigated.}
% First, the same experimental setting is performed with SKD in figurex. We can see that the KD loss is largely reduced, demonstrating the underfitting problem is highly mitigated.

This can be credited to that SKD is easy to optimize. As Section 3 and Section 4 discussed, KD let the \emph{student} learn both norm and normalized logits from teacher, while SKD focus on the normalized logits, which is more effective than the logits itself.

\begin{figure}[t]
\begin{center}
   \includegraphics[width=0.95\linewidth]{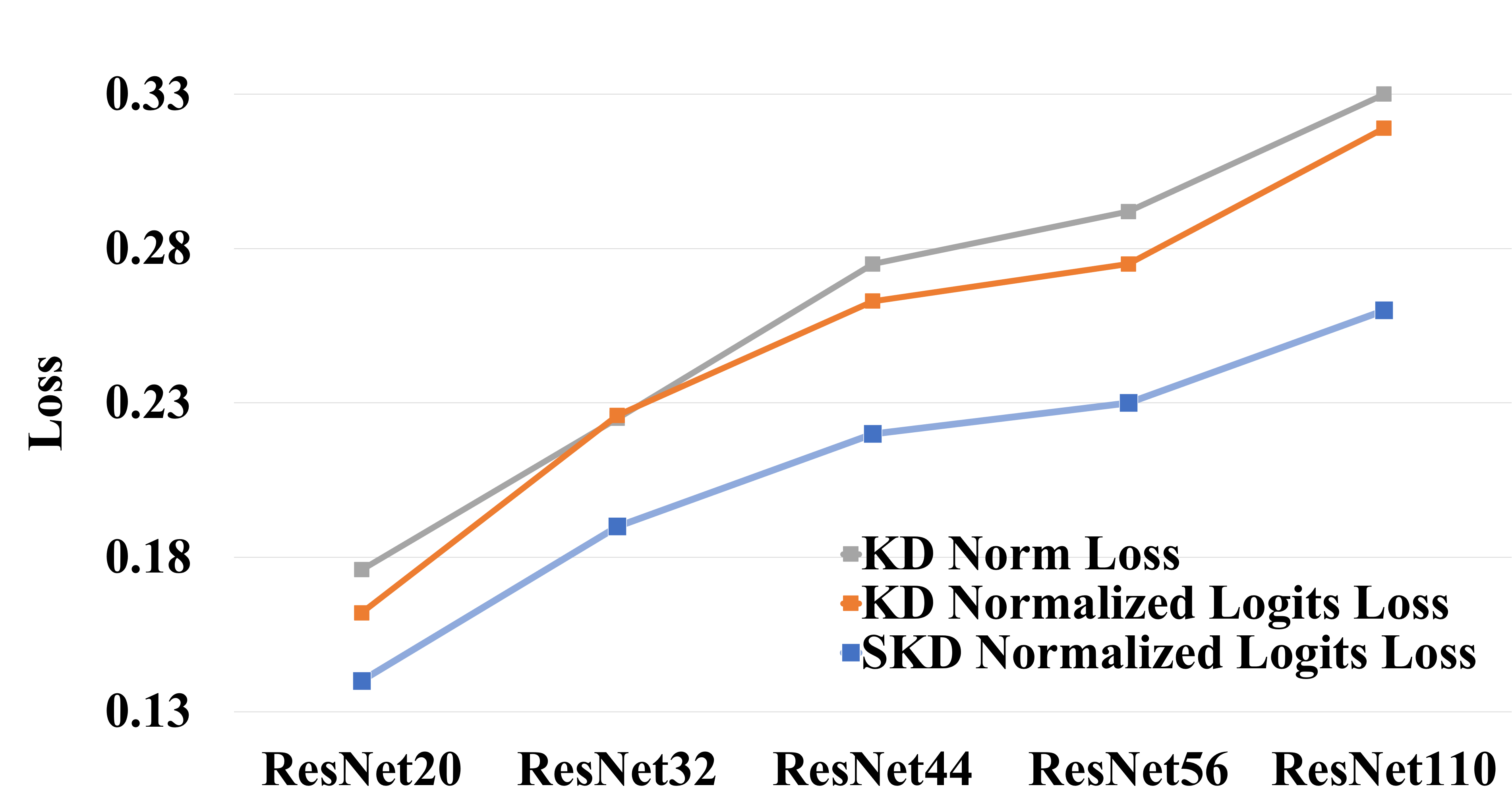}
\end{center}
   \caption{The norm and normalized logits difference (MSE Loss) between \emph{teacher} and \emph{student} at the end of training, with KD or SKD. The normalized logits difference of SKD is much lower than KD. SKD does not learn the norm, which gives it more capacity to focus on learning the normalized logits, which is more useful to logits (Section \ref{norm_and_normalized_logits}).}
\label{fig:capacity_loss}
\end{figure}

We conducted experiments to evaluate how \emph{students} fit these two parts in practice, and we find SKD fitting the normalized logits much better than KD. We log the MSE loss of the norm and normalized logits between the \emph{teacher} and \emph{student} at the end of the training. In Fig. \ref{fig:capacity_loss}, show the norm loss and normalized losses. The horizontal plot with the \emph{teacher} size, while the vertical varies the value of the loss. In the KD experiment, we can see that the \emph{student} faces problems to fit both the norm and normalized logits. When the \emph{teacher} grows bigger, both losses increase. 

On the other hand, the norm loss is zero under SKD, and the fitting of normalized logits is also eased by a large margin compared to the KD. SKD eliminates norm learning, which give it more capacity to focus on learning the normalized logits.

\begin{figure}[t]
\begin{center}
   \includegraphics[width=0.95\linewidth]{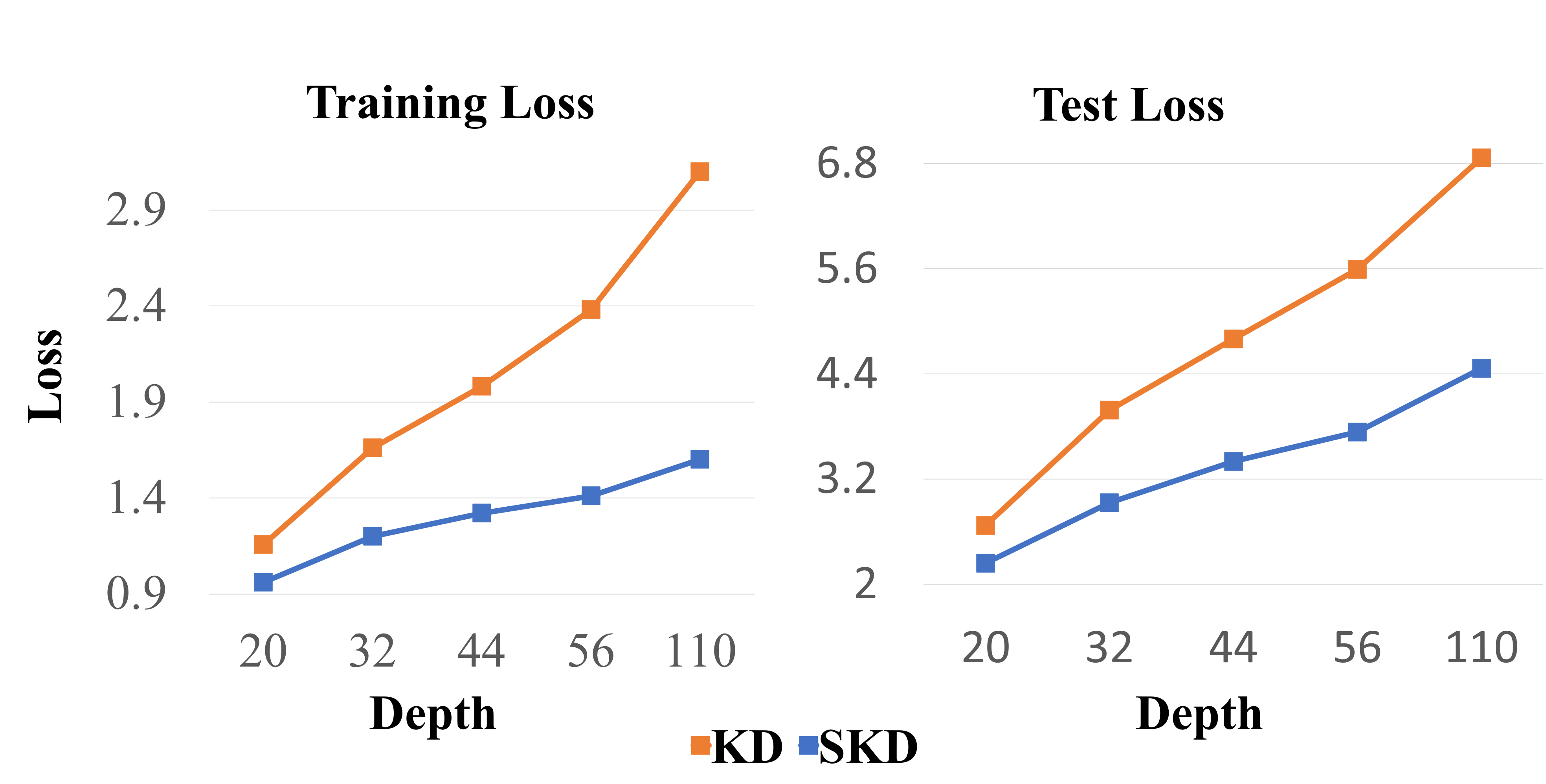}
\end{center}
   \caption{SKD has much lower training/test loss. With the \emph{teacher} gets larger, the loss increase slower than KD, shows that SKD alleviates the Capacity Gap Problem.}
\label{fig:kd_loss}
\end{figure}

Furthermore, in Fig. \ref{fig:kd_loss} we show the training/test loss with different teachers. SKD has much lower loss compared to KD when the teacher gets larger.

\subsection{SKD is Robust to Temperature}

Temperature is an important hyper-parameter in knowledge distillation to soften the output probability distribution. One might want to use a higher temperature to soften the output distribution to alleviate the capacity gap problem. There has been some precedent researches~\cite{Earlystop, HintonKD} that did not support this method. 

% The reason behind this result is still under-explored. In this chapter, we revisit this problem and explore the relationships between temperature and the Capacity Gap problem.

\begin{figure}[t]
\centering
\centering
\includegraphics[width=0.99\linewidth]{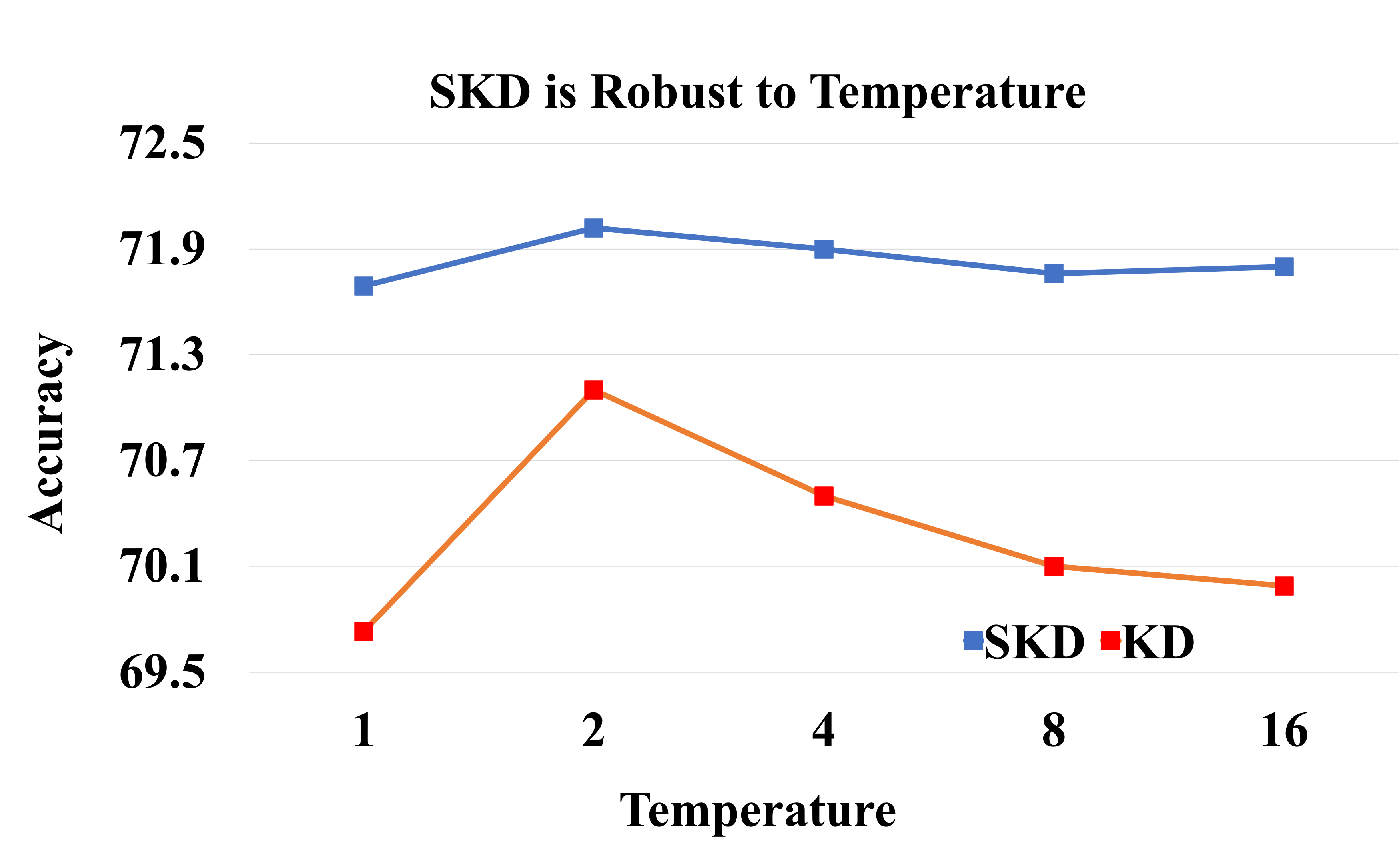}
\caption{Experiments with a broad range of temperatures. SKD is very robust to temperatures. However, KD must hand pick a narrow range of temperatures to train students. SKD is better than KD in all settings.}
\label{temp}
\end{figure}

Here we use ResNet56 as~\emph{teacher} and ResNet20 as~\emph{student} to evaluate the performance of KD and SKD with the multiple temperatures.
The temperature varies from a broad range. Fig.~\ref{temp} shows both the results of Hinton KD and SKD. From the results, we can find that the performance of knowledge distillation is very sensitive to temperature. As the temperature increases, the performance of~\emph{student} first increases and then decreases by a large range. Thus as for Hinton KD, we need to tune this parameter to get the best performance carefully, and demonstrates that higher temperature cannot alleviate the capacity gap problem. On the other side, the performance of SKD is robust to temperature, which achieves excellent performance with all temperature settings. 

\section{Discussion}
A large model can produce a very peaky distribution to fit the discrete ground truth during the training process, where a small can only shape a soft distribution. Indeed, a model is limited to its capacity capability, that some objective goal is beyond its reach. For example, the well known XOR problem is impossible for a perceptron to fit. So to the small model to produce a peaky output to fit the discrete distribution.

% to fit the discrete example-label pairs, a large model can learn a sharp distribution
Under this perspective, a soft target produced by another bigger model is a more appropriate goal for the small model to learn. After all, this target is softer than the ground truth. 
Furthermore, the peaky nature of such a soft target is still a challenging objective. A more wise move would be to free the small model from entirely fitting this overconfidence. This is where SKD steps in.  SKD considers the limited representation capability of compact models to ease the training and achieve excellent performance.

% The limited capacity is the critical difference between training a cumbersome or a compact model. 

% It is widely known that some complex mapping functions are beyond reach for certain neural networks with limited capacity. 

% , while in knowledge distillation, the discrete distribution is challenging for this small model to fit.

{\small
\bibliographystyle{ieee_fullname}
\bibliography{egbib}
}

\end{document}